\documentclass{article}

\usepackage[hidelinks]{hyperref}
\usepackage{spconf,amsmath,graphicx}
\usepackage{booktabs,multirow}
\usepackage[misc]{ifsym}

\pdfoutput=1

\makeatletter
\renewcommand{\name}[1]{\gdef\@name{\em #1}}
\makeatother

\title{RationAnomaly: Log Anomaly Detection with Rationality via Chain-of-Thought and Reinforcement Learning}

\name{Song Xu$^{*\star\dagger}$, Yilun Liu$^{\dagger}$\textsuperscript{\Letter}, Minggui He$^{\dagger}$, Mingchen Dai$^{*\star\dagger}$, Ziang Chen$^{\S\dagger}$, \\
\textit{Chunguang Zhao$^{\dagger}$, Jingzhou Du$^{\dagger}$, Shimin Tao$^{\dagger}$, Weibin Meng$^{\dagger}$,}\\
\textit{Shenglin Zhang$^{\S}$, Yongqian Sun$^{\S}$, Boxing Chen$^{\ddagger}$, Daimeng Wei$^{\dagger}$} \\ \thanks{\textsuperscript{\Letter} Corresponding author (liuyilun3@huawei.com).}}
\address{$^{*}$School of Software Engineering, University of Science and Technology of China, Hefei, China\\
    $^{\star}$Suzhou Institute for Advanced Research, University of Science and Technology of China, Suzhou, China\\
    $^{\dagger}$Huawei, Beijing, China\\$^{\ddagger}$Huawei Canada, Montreal, Canada\\$^{\S}$Nankai University, Tianjin, China}

\begin{document}

\maketitle

\begin{abstract}
Logs constitute a form of evidence signaling the operational status of software systems. Automated log anomaly detection is crucial for ensuring the reliability of modern software systems. However, existing approaches face significant limitations: traditional deep learning models lack interpretability and generalization, while methods leveraging Large Language Models are often hindered by unreliability and factual inaccuracies. To address these issues, we propose RationAnomaly, a novel framework that enhances log anomaly detection by synergizing Chain-of-Thought (CoT) fine-tuning with reinforcement learning. Our approach first instills expert-like reasoning patterns using CoT-guided supervised fine-tuning, grounded in a high-quality dataset corrected through a rigorous expert-driven process. Subsequently, a reinforcement learning phase with a multi-faceted reward function optimizes for accuracy and logical consistency, effectively mitigating hallucinations. Experimentally, RationAnomaly outperforms state-of-the-art baselines, achieving superior F1-scores on key benchmarks while providing transparent, step-by-step analytical outputs. We have released the corresponding resources, including code and datasets\footnote{https://github.com/Gravityless/RationAnomaly}.
\end{abstract}


\section{Introduction}
\label{sec:introduction}

Logs represent a type of signal generated by software systems to attest to their operational state. Automated log analysis through machine learning has become essential for maintaining system reliability \cite{backgd1,backgd2}. However, existing approaches face significant limitations: traditional deep learning models lack interpretability and generalize poorly \cite{howfar}, while LLM-based methods are hindered by unreliability and factual inaccuracies \cite{backgd3}. Prompt-based LLM approaches suffer from hallucinations \cite{loglm}, and existing fine-tuning strategies fail to explicitly model step-by-step diagnostic reasoning \cite{loggpt}.

To overcome the limitations, we introduce RationAnomaly, a novel framework that integrates Chain-of-Thought fine-tuning with reinforcement learning. Our approach begins with CoT-guided supervised fine-tuning on a high-quality, expert-validated dataset to instill systematic reasoning patterns. This is followed by a reinforcement learning phase employing a multi-faceted reward function that optimizes both detection accuracy and logical consistency, thereby effectively reducing hallucinations. Experiments demonstrate that RationAnomaly achieves state-of-the-art performance, delivering superior accuracy while offering transparent and interpretable analytical outputs.

\begin{figure}[htb]
\centering
\centerline{\includegraphics[width=0.43\textwidth]{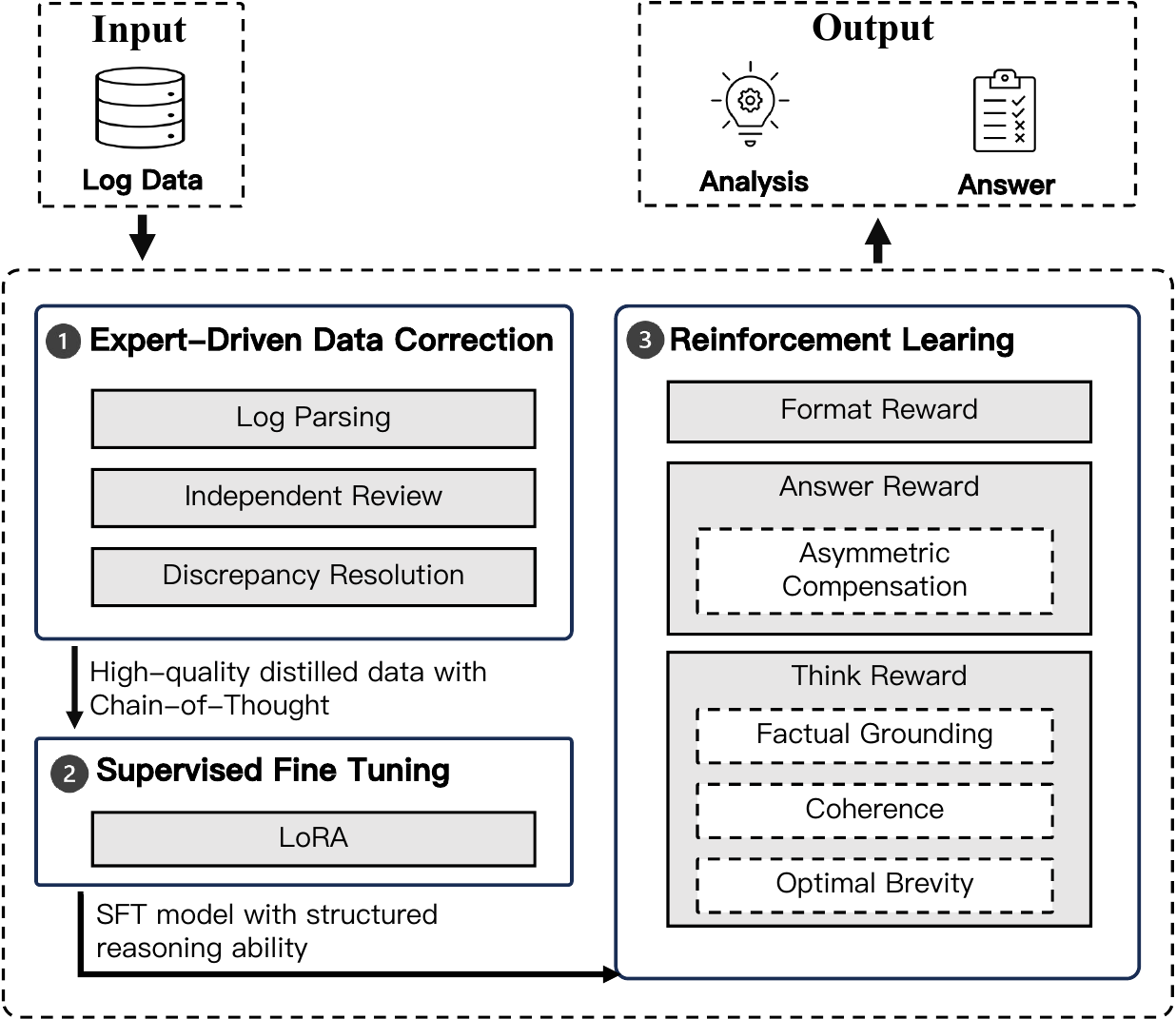}}
\caption{Overview of RationAnomaly}
\label{fig:overview}
\end{figure}

\section{RELATED WORK}
\label{sec:relatedwork}

\subsection{Deep Learning-based Log Anomaly Detection}
There is a significant amount of research on applying deep learning to log anomaly detection \cite{logsy,plelog,swisslog,logflash,logrep}. DeepLog \cite{deeplog} used LSTMs to model normal patterns, while subsequent approaches enhanced performance with template embeddings (LogAnomaly \cite{loganomaly}) and attention mechanisms (LogRobust \cite{logrobust}). Despite achieving high accuracy, these methods operate as black boxes, lacking interpretability.

\subsection{LLM-based Log Anomaly Detection}
Recent research leverages LLMs through prompt-based methods and fine-tuning approaches. Prompt-based methods include LogPrompt \cite{logprompt}, EagerLog \cite{eagerlog}, LogRAG \cite{lograg} and RAGLog \cite{raglog}, while fine-tuning methods include LogGPT \cite{loggpt}, LogLM \cite{loglm} and SuperLog \cite{superlog}. These approaches either suffer from hallucinations or fail to explicitly model step-by-step reasoning.

Our work addresses this gap by explicitly optimizing the reasoning process through a two-stage paradigm: CoT-guided fine-tuning followed by reinforcement learning alignment, uniquely combining accuracy with interpretability.

\section{METHODOLOGY}
\label{sec:methodology}

Our framework, RationAnomaly, enhances log anomaly detection through a multi-stage process, as illustrated in Fig.~\ref{fig:overview}. The process begins with a foundational data correction stage to ensure data integrity, followed by a two-stage training paradigm: (1) \textbf{Chain-of-Thought Supervised Fine-Tuning (CoT-SFT)} to instill expert-like reasoning, and (2) \textbf{Reinforcement Learning Alignment (RLA)} to optimize for accuracy and reliability.

\begin{table*}[t]
\centering
\caption{Detailed breakdown of annotation errors identified and corrected in the dataset.}
\label{tab:corrected_logs}
\vspace{0.5em}
\begin{tabular}{l|c|c|c|p{6.5cm}}
\toprule
\textbf{Correction} & \textbf{Error Category} & \textbf{Count} & \textbf{Percentage} & \textbf{Representative Examples} \\
\midrule
\multirow{12}{*}{Normal → Abnormal} & System Error & 78 & 34.7\% & ``PANIC: segmentation violation...'' \\
& & & & ``hit ASSERT condition...'' \\
\cmidrule{2-5}
& Network Issue & 47 & 20.9\% & ``Connection refused...'' \\
& & & & ``Bad UMNT RPC: RPC: Timed out'' \\
\cmidrule{2-5}
& Hardware Failure & 40 & 17.8\% & ``parity error in read queue...'' \\
& & & & ``DDR failing info register...'' \\
\cmidrule{2-5}
& Software Exception & 56 & 24.9\% & ``ciod: Error loading...'' \\
& & & & ``divide-by-zero...'' \\
\midrule
Abnormal → Normal & - & 4 & 1.8\% & ``exited normally with exit code...'' \\
& & & & ``Mounting NFS filesystems...'' \\
\midrule
\multicolumn{1}{c|}{\bfseries Total} & \multicolumn{1}{c|}{\bfseries -} & \multicolumn{1}{c|}{\bfseries 225} & \multicolumn{1}{c|}{\bfseries 100.0\%} & \multicolumn{1}{c}{\bfseries -} \\
\bottomrule
\end{tabular}
\end{table*}

\subsection{Expert-Driven Data Correction}

The reliability of any data-driven model is contingent upon the quality of its training and evaluation data. Public benchmarks like BGL and Spirit, while widely used \cite{loghub}, contain systematic labeling errors that can compromise model evaluation. To establish a trustworthy foundation, we conducted a comprehensive data correction process.

We assembled a team of five industry experts to systematically review all 3,046 unique log templates extracted from the BGL and Spirit datasets. The process began with an independent review by each expert. Discrepancies were then resolved through consensus-driven panel discussions, with a senior expert providing final validation for disputed cases. As shown in Table~\ref{tab:corrected_logs}, this rigorous process identified and corrected 225 systematically mislabeled log templates (7.4\% of total). Our analysis revealed a significant bias towards false negatives (98.2\% of errors), where critical system failures (e.g., ``segmentation violation'', ``Connection refused'') were incorrectly marked as normal. The result is a high-fidelity benchmark characterized by inter-annotator agreement at $\kappa$=0.94, forming a reliable basis for our subsequent training and evaluation.

\subsection{Chain-of-Thought Supervised Fine-Tuning}

The goal of this stage is to imbue the model with structured, reasoning capabilities before it delivers a final verdict. To achieve this, we leverage a powerful teacher model to distill a CoT dataset. For each log template in the training set, we prompted the teacher model to generate a step-by-step analysis that mimics an expert's diagnostic process: identifying key parameters, reasoning about their implications, and drawing a conclusion. This yields a high-quality dataset of (log, CoT-analysis, label) triplets. To maintain computational efficiency and prevent catastrophic forgetting of the base model's capabilities, we use Low-Rank Adaptation (LoRA) \cite{lora}.

\subsection{Reinforcement Learning Alignment}

While SFT teaches reasoning patterns, it does not guarantee factual accuracy or reliability against hallucinations when it comes to various logs. The RLA stage addresses this by refining the model's behavior, aligning it with real-world operational goals through a meticulously designed reward function.

We employ Group Relative Policy Optimization (GRPO), an efficient and stable RL algorithm, to optimize the model. The cornerstone of this stage is our multi-faceted reward function, $R_{total}$, which provides a holistic evaluation of the model's generated output from three perspectives: format adherence, answer correctness, and reasoning quality.

\textbf{Format Reward} ($R_{format}$): A binary reward that is positive only if the output strictly adheres to the predefined structure (i.e., contains both $\langle$think$\rangle$ and $\langle$answer$\rangle$ sections). A failure here will result in the subsequent rewards being ignored, strongly discouraging malformed outputs.

\textbf{Answer Reward} ($R_{answer}$): To address the high cost of false negatives in production environments, and the unbalanced distribution between normal and abnormal classes in the dataset, we introduce an asymmetric reward mechanism: correctly identifying an anomaly results in a higher reward than correctly identifying a normal log, and missing an anomaly also leads to a heavier penalty.

\textbf{Thinking Reward} ($R_{think}$): To combat hallucination, the model's output is optimized along three crucial dimensions:

(1) Factual Grounding: Assesses the semantic overlap (using BLEU and ROUGE) between the generated analysis and the source content. This dimension ensures the reasoning is directly supported by evidence from the log, effectively discouraging hallucination.

(2) Coherence: Utilizes a perplexity model to evaluate the fluency and logical flow of the reasoning. This promotes outputs that are sensible and easy to follow, rather than repetitive or nonsensical.

(3) Optimal Brevity: Encourages concise yet complete explanations by assessing alignment with the target length derived from our distilled CoT dataset. This ensures the analysis is efficient without sacrificing critical information.

\begin{table*}[t]
\centering
\caption{Overall Result}
\label{tab:overall_result}
\vspace{0.5em}
\begin{tabular}{c ccc ccc ccc ccc}
\toprule
\multirow{2}{*}{\textbf{Method}} & \multicolumn{3}{c}{\textbf{BGL(Session-level)}} & \multicolumn{3}{c}{\textbf{BGL(Template-level)}} & \multicolumn{3}{c}{\textbf{Spirit(Session-level)}} & \multicolumn{3}{c}{\textbf{Spirit(Template-level)}} \\
\cmidrule(lr){2-4} \cmidrule(lr){5-7} \cmidrule(lr){8-10} \cmidrule(lr){11-13}
& F1 & Pre & Rec & F1 & Pre & Rec & F1 & Pre & Rec & F1 & Pre & Rec \\
\midrule
DeepLog               & 0.869  & 0.811  & 0.935  & -      & -      & -      & 0.905  & 0.891  & 0.919  & -      & -      & -        \\
LogAnomaly            & 0.857  & 0.770  & \textbf{0.965}  & -      & -      & -      & 0.925  & 0.880  & 0.975  & -      & -      & -        \\
LogRobust             & 0.811  & 0.704  & 0.956  & -      & -      & -      & 0.921  & 0.856  & \textbf{0.996}  & -      & -      & -        \\
LogPrompt             & 0.834  & 0.874  & 0.811  & 0.827  & 0.724  & \textbf{0.964}  & 0.917  & 0.859  & 0.983  & 0.795  & 0.834  & \textbf{0.858}    \\
Llama 2 7B            & 0.707  & 0.741  & 0.753  & 0.686  & 0.858  & 0.633  & 0.800  & 0.881  & 0.861  & 0.655  & 0.839  & 0.598    \\
RationAnomaly          & \textbf{0.909}  & \textbf{0.900}  & 0.919  & \textbf{0.887}  & \textbf{0.898}  & 0.881  & \textbf{0.958}  & \textbf{0.959}  & 0.959  & \textbf{0.862}  & \textbf{0.899}  & 0.847    \\
\bottomrule
\end{tabular}
\end{table*}

\section{EXPERIMENTS}
\label{sec:experiments}

\subsection{Experimental Setup}

For the BGL and Spirit datasets, we performed a chronological split to simulate real-world data flow. To support deep learning baselines, we first sampled a 2000-log template-level training set (15\% anomaly rate), then constructed session-level data using a 100-log fixed window \cite{howfar}. Test sets were created similarly, resulting in 8000 entries and sessions per dataset. To prevent data leakage, logs used in RationAnomaly's training were excluded from these session-level test sets.

We compare RationAnomaly against two categories of state-of-the-art methods: conventional deep learning models, including unsupervised approaches like DeepLog \cite{deeplog} and LogAnomaly \cite{loganomaly} as well as the supervised LogRobust \cite{logrobust}; and LLM-based techniques, such as LogPrompt \cite{logprompt} and a zero-shot Llama 2 7B baseline.

RationAnomaly is built upon Llama 2 7B and implemented using PyTorch, VeRL and Hugging Face libraries. We use three standard metrics for evaluation: Precision, Recall, and F1-score.

\subsection{Overall Performance}

Table~\ref{tab:overall_result} presents a comprehensive performance comparison. RationAnomaly establishes a new state-of-the-art, achieving superior F1-scores across all evaluated datasets and scenarios.

Our method's performance surpasses that of conventional deep learning baselines. On session-level, \textbf{RationAnomaly achieves an F1-score of 0.958} on Spirit, a notable improvement over the best-performing baseline, LogAnomaly (0.925). Furthermore, unlike traditional methods which are limited to session-level analysis, RationAnomaly's semantic understanding enables effective template-level detection, achieving F1-scores of 0.887 on BGL and 0.862 on Spirit. This demonstrates its advanced capability for fine-grained log interpretation.

The performance delta between RationAnomaly and the zero-shot Llama 2 7B baseline highlights the necessity of our two-stage training paradigm. On the BGL template-level, \textbf{the F1-score shows a 29.3\% relative improvement}. Moreover, a critical outcome of our method is the achievement of a \textbf{well-balanced precision-recall profile}. On the Spirit session dataset, it attains 0.959 for both precision and recall. This balance is a direct consequence of the Reinforcement Learning Alignment stage, which trains the model to be both accurate in its predictions and sensitive to genuine anomalies, enhancing its reliability for operational use.

\begin{figure}[t]
\centering
\includegraphics[width=0.48\textwidth]{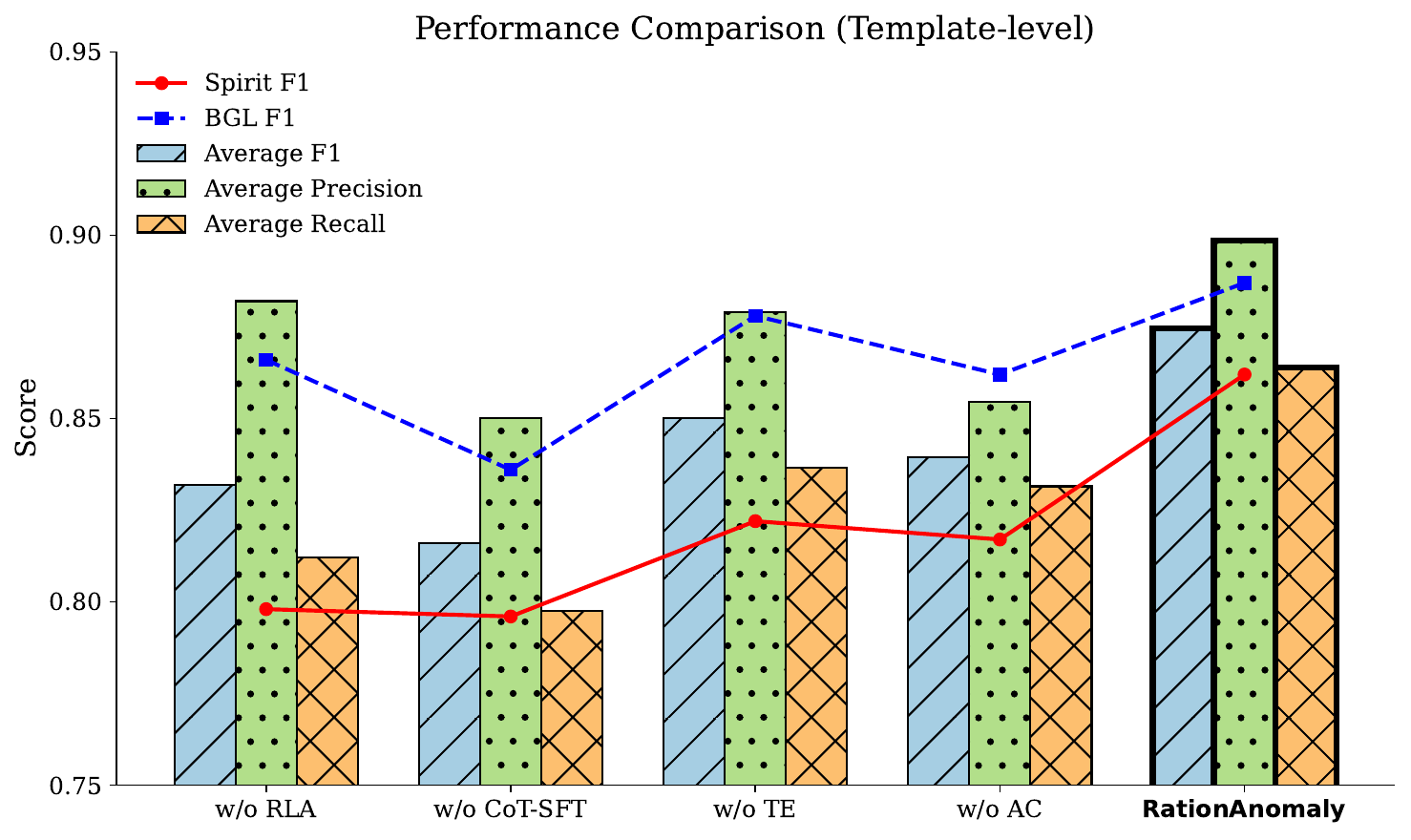}
\caption{Ablation Result}
\label{fig:ablation_result}
\end{figure}

\subsection{Ablation Study}

To dissect the contribution of each component within our framework, we conducted an ablation study. As shown in Fig.~\ref{fig:ablation_result}, the ablation experiment results show the same trend on BGL and Spirit datasets.

The model without CoT-SFT solely undergoes reinforcement learning achieves the lowest F1-score and significantly lower Precision and Recall values. This indicates that fine-tuning on a structured reasoning process provides an important and effective foundation for subsequent steps. 

The necessity of RLA stage is made evident by the significant performance increase from the model without RLA to the final RationAnomaly model, which is reflected in \textbf{F1-score improvement from 0.798 to 0.862} on Spirit. This result indicates that RLA is an essential step for refining the model's behavior, correcting subtle reasoning flaws, and aligning its outputs toward maximal accuracy.

The specific design of the reward function proves to be pivotal. Disabling the Asymmetric Compensation (without AC) or the Thinking Evaluation (without TE) leads to distinct performance degradation, with \textbf{F1-scores dropping from 0.862 to 0.817 and 0.822} on Spirit. This confirms that both applying greater penalties to high-cost errors and rewarding factually-grounded reasoning are critical mechanisms for achieving the model's final, robust performance.

\begin{figure}[t]
\centering
\includegraphics[width=0.48\textwidth]{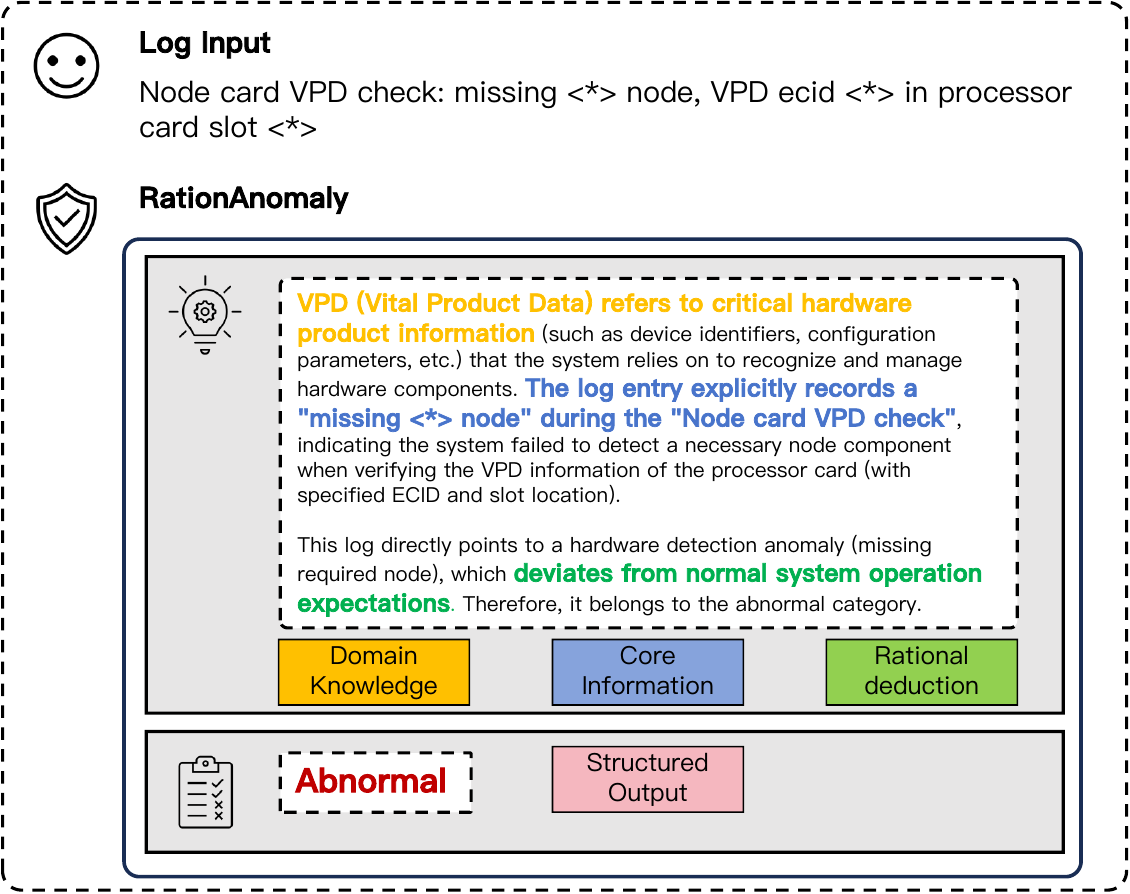}
\caption{A case study demonstrating RationAnomaly's step-by-step reasoning on a hardware-related log.}
\label{fig:case_study}
\end{figure}

\subsection{Case Study}

To demonstrate our model's capabilities, we present a case study in Fig.~\ref{fig:case_study}. The log entry requires domain-specific knowledge that challenges conventional methods. RationAnomaly not only correctly classifies it as abnormal but also generates a transparent, expert-like rationale. It leverages its acquired \textbf{Domain Knowledge} to define VPD, extracts the \textbf{Core Information} (``missing node''), and performs a \textbf{Rational Deduction} to identify a hardware detection failure. This structured output, a direct result of our CoT-SFT and RLA pipeline, showcases a shift from simple pattern matching to providing verifiable and trustworthy diagnostic insights.

\section{CONCLUSION}
\label{sec:conclusion}

In this paper, we introduced RationAnomaly, a novel framework that significantly enhances the reliability and interpretability of log anomaly detection. Our two-stage paradigm, grounded in expert-corrected data, synergizes Chain-of-Thought fine-tuning with reinforcement learning alignment. Experiments demonstrate that RationAnomaly substantially outperforms existing methods, achieving superior accuracy while providing transparent, step-by-step reasoning. This work represents a significant step towards developing AIOps tools that are dependable and trustworthy, with future possibilities for extension into correlating multi-modal signals from logs, metrics, and traces.

\vfill\pagebreak

\bibliographystyle{IEEEbib}
\bibliography{refs}

\end{document}